\documentclass[10pt,twocolumn,letterpaper]{article}

\usepackage{cvpr}
\usepackage{times}
\usepackage{epsfig}
\usepackage{graphicx}
\usepackage{amsmath}
\usepackage{amssymb}

\usepackage[usenames, dvipsnames]{color} 
\usepackage{verbatim} 
\usepackage{tabularx} 
\usepackage{multirow} 
\usepackage{float} 
\usepackage{mathtools} 
\usepackage[normalem]{ulem} 
\usepackage{caption} 

\definecolor{purple}{rgb}{0.5, 0.0, 0.5}
\definecolor{lightgreen}{rgb}{0.68, 1, 0.18}
\definecolor{darkgreen}{rgb}{0.09, 0.32, 0.24}
\definecolor{darkred}{rgb}{0.6, 0, 0}
\iftrue 
  \newcommand{\vitto}[1]{\textcolor{red}{\bf [VF: #1]}}
  \newcommand{\holgerold}[1]{\textcolor{lightgreen}{\bf [HC: #1]}}
  \newcommand{\holgernew}[1]{\textcolor{red}{\bf [HC: #1]}}
  \newcommand{\jasper}[1]{\textcolor{blue}{\bf [JU: #1]}}
  \newcommand{\done}[1]{\textcolor{cyan}{\bf [Done: #1]}}
  \newcommand{\action}[1]{\textcolor{red}{\bf [Action: #1]}}
  \newcommand{\todo}[1]{\textcolor{red}{\bf [Todo: #1]}}
\else
  \newcommand{\vitto}[1]{\noindent}
  \newcommand{\holgerold}[1]{\noindent}
  \newcommand{\holgernew}[1]{\noindent}
  \newcommand{\jasper}[1]{\noindent}
  \newcommand{\done}[1]{\noindent}
  \newcommand{\action}[1]{\noindent}
  \newcommand{\todo}[1]{\noindent}
\fi
\newcommand{\mypartop}[1]{\vspace{0mm}\paragraph{#1}}
\newcommand{\mypar}[1]{\vspace{-5mm}\paragraph{#1}}
\newcommand{\presectionspace}{\vspace{-3mm}}
\newcommand{\presubsectionspace}{\vspace{-1mm}}
\newcommand{\figcapstartspace}{\vspace{-5mm}}
\newcommand{\figcapendspace}{\vspace{-4mm}}
\newcommand{\tabcapstartspace}{\vspace{-2mm}}
\newcommand{\tabcapendspace}{\vspace{-5mm}}
\newcommand{\tabresize}[1]{\resizebox{1\linewidth}{!}{#1}}

\newcolumntype{C}[1]{>{\centering\let\newline\\\arraybackslash\hspace{0pt}}m{#1}} 
\newcolumntype{L}[1]{>{\let\newline\\\arraybackslash\hspace{0pt}}m{#1}} 
\newcolumntype{R}[1]{>{\raggedleft\let\newline\\\arraybackslash\hspace{0pt}}m{#1}} 
\makeatletter

\usepackage[colorlinks=true,linkcolor=green]{hyperref}

\cvprfinalcopy 

\def\cvprPaperID{735} 

\ifcvprfinal\fi

\begin{document}

\title{COCO-Stuff: Thing and Stuff Classes in Context}

\author{
Holger Caesar\textsuperscript{1} \; \; 
Jasper Uijlings\textsuperscript{2} \; \; 
Vittorio Ferrari\textsuperscript{1\,2}\\
University of Edinburgh\textsuperscript{1} \; \, 
Google AI Perception\textsuperscript{2}
}

\maketitle

\begin{abstract}
Semantic classes can be either things (objects with a well-defined shape, e.g. car, person) or stuff (amorphous background regions, e.g. grass, sky).
While lots of classification and detection works focus on thing classes, less attention has been given to stuff classes.
Nonetheless, stuff classes are important as they allow to explain important aspects of an image, including
(1) scene type;
(2) which thing classes are likely to be present and their location (through contextual reasoning);
(3) physical attributes, material types and geometric properties of the scene.
To understand stuff and things in context we introduce COCO-Stuff\,\footnote{\url{http://calvin.inf.ed.ac.uk/datasets/coco-stuff}\label{dataseturl}},
which augments all 164K images of the COCO 2017 dataset with pixel-wise annotations for 91 stuff classes.
We introduce an efficient stuff annotation protocol based on superpixels, which leverages the original thing annotations.
We quantify the speed versus quality trade-off of our protocol and explore the relation between annotation time and boundary complexity. 
Furthermore, we use COCO-Stuff to analyze: 
(a) the importance of stuff and thing classes in terms of their surface cover and how frequently they are mentioned in image captions;
(b) the spatial relations between stuff and things, highlighting the rich contextual relations that make our dataset unique;
(c) the performance of a modern semantic segmentation method on stuff and thing classes, and whether stuff is easier to segment than things.
\end{abstract}

\presectionspace
\vspace{-7mm}
\section{Introduction}

Most of the recent object detection efforts have focused on recognizing and localizing thing classes, such as \emph{cat} and \emph{car}.
Such classes have a specific size~\cite{forsyth96orcv,heitz08eccv} and shape~\cite{forsyth96orcv,tighe13cvpr,uijlings13ijcv,mottaghi13cvpr,endres14pami,dai15cvpr}, 
and identifiable parts (e.g. a \emph{car} has \emph{wheels}).
Indeed, the main recognition challenges~\cite{everingham15ijcv,russakovsky15ijcv,lin14eccv} are all about things.
In contrast, much less attention has been given to stuff classes, such as \emph{grass} and \emph{sky},
which are amorphous and have no distinct parts (e.g. a piece of \emph{grass} is still \emph{grass}).
In this paper we ask: Is this strong focus on things justified?

To appreciate the importance of stuff, consider that it makes up the majority of our visual surroundings.
For example, \emph{sky}, \emph{walls} and most \emph{ground} types are stuff.
Furthermore, stuff often determines the type of a scene,
so it can be very descriptive for an image (e.g. in a beach scene the \emph{beach} and \emph{water} are the essential elements, more so than \emph{people} and \emph{volleyball}).
Stuff is also crucial for reasoning about things:
Stuff captures the 3D layout of the scene and therefore heavily constrains the possible locations of things.
The contact points between stuff and things are critical for determining depth ordering and relative positions of things,
which supports understanding the relations between them.
Finally, stuff provides context helping to recognize small or uncommon things,
e.g. a metal thing in the sky is likely an \emph{aeroplane}, while a metal thing in the water is likely a \emph{boat}.
For all these reasons, stuff plays an important role in scene understanding and we feel it deserves more attention.

\begin{figure}
\small
\centering
\includegraphics[width=\linewidth]{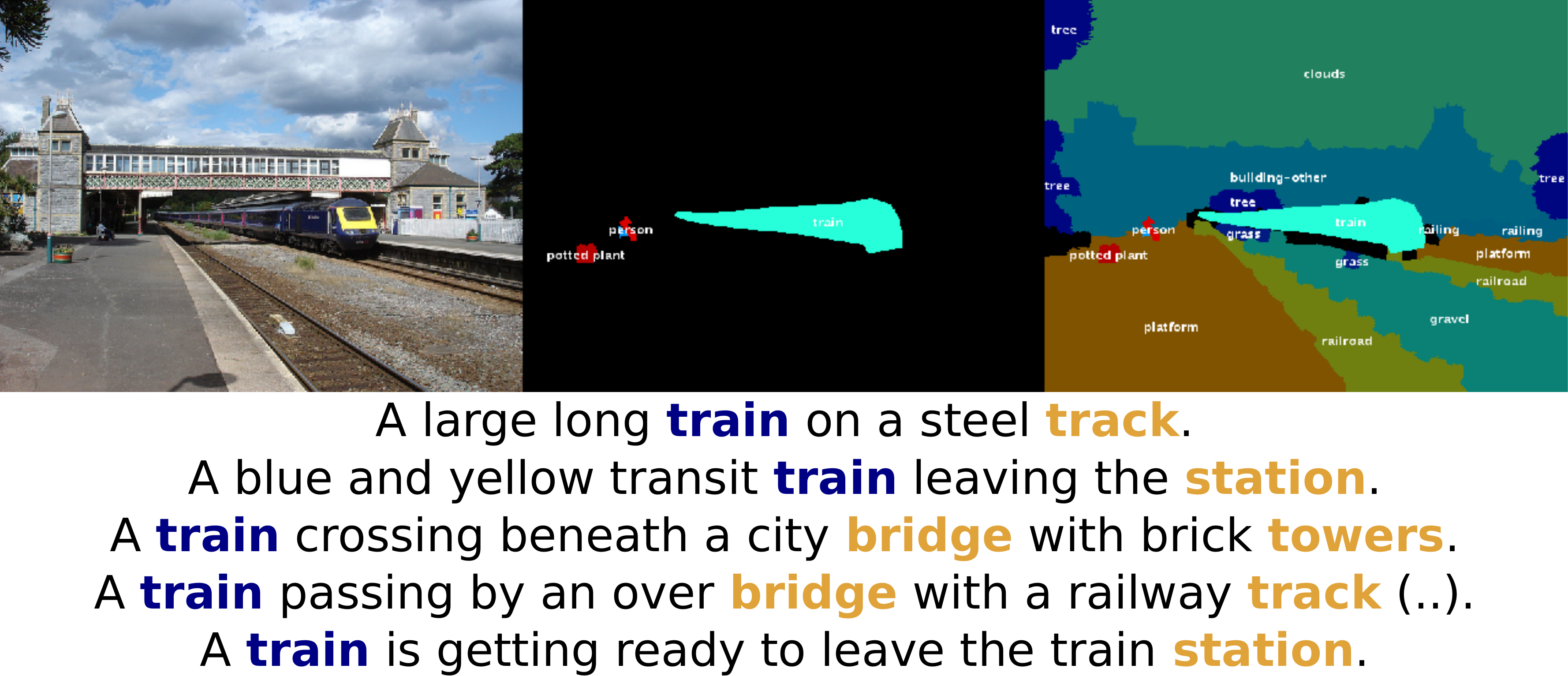}
\figcapstartspace
\vspace{-2mm}
\caption{
\textit{
\small
(left) An example image, (middle) its thing annotations in COCO~\cite{lin14eccv} and (right) enriched stuff and thing annotations in COCO-Stuff.
Just having the \emph{train}, \emph{person}, \emph{bench} and \emph{potted plant} does not tell us much about the context of the scene,
but with stuff and thing labels we can infer the position and orientation of the \emph{train}, its stuff-thing interactions~(\emph{train} leaving the \emph{station})
and thing-thing interactions~(\emph{person} waiting for a different \emph{train}).
This is also visible in the captions written by humans.
Whereas the captions only mention one thing (\emph{train}), they describe a multitude of different stuff classes~(\emph{track}, \emph{station}, \emph{bridge}, \emph{tower}, \emph{railway}),
stuff-thing interactions~(\emph{train} leaving the \emph{station}, \emph{train} crossing beneath a city \emph{bridge}) and spatial arrangements~(on, beneath).
}
}
\figcapendspace
\vspace{-3mm}
\label{fig:introexample}
\end{figure}

In this paper we introduce the COCO-Stuff dataset, which augments the popular COCO~\cite{lin14eccv} with pixel-wise annotations for a rich and diverse set of 91 stuff classes.
The original COCO dataset already provides outline-level annotation for 80 thing classes.
The additional stuff annotations enable the study of stuff-thing interactions in the complex COCO images.
To illustrate the added value of our stuff annotations, Fig.~\ref{fig:introexample} shows an example image, its annotations in COCO and COCO-Stuff.
The original COCO dataset offers location annotations only for the \emph{train}, \emph{potted plant}, \emph{bench} and \emph{person},
which are not sufficient to understand what the scene is about.
Indeed, the image captions written by humans (also provided by COCO) mention the \emph{train}, its interaction with stuff (i.e. \emph{track}),
and the spatial arrangements of the \emph{train} and its surrounding stuff.
All these elements are necessary for scene understanding and show how COCO-Stuff offers much more comprehensive annotations.

This paper makes the following contributions:
(1) We introduce COCO-Stuff, which augments the original COCO dataset with stuff annotations. 
(2) We introduce an annotation protocol for COCO-Stuff which leverages the existing thing annotations and superpixels. We demonstrate both the quality and efficiency of this protocol (Sec.~\ref{sec:dataset}).
(3) Using COCO-Stuff, we analyze the role of stuff from multiple angles (Sec.~\ref{sec:analysis}):
(a) the importance of stuff and thing classes in terms of their surface cover and how frequently they are mentioned in image captions;
(b) the spatial relations between stuff and things, highlighting the rich contextual relations that make COCO-Stuff unique;
(c) we compare the performance of a modern semantic segmentation method on thing and stuff classes.

Hoping to further promote research on stuff and stuff-thing contextual relations,
we release COCO-Stuff and the trained segmentation models \href{https://github.com/nightrome/cocostuff}{online}\textsuperscript{\ref{dataseturl}}.
\presectionspace
\vspace{+0mm}
\section{Related Work}
\label{sec:relatedwork}
\vspace{-0mm}
\mypartop{Defining things and stuff.}
The literature provides definitions for several aspects of stuff and things, including:
(1) Shape:
Things have characteristic shapes (\emph{car}, \emph{cat}, \emph{phone}),
whereas stuff is amorphous (\emph{sky}, \emph{grass}, \emph{water}) \cite{forsyth96orcv,xiao10cvpr,ion11nips,tighe13cvpr,uijlings13ijcv,mottaghi13cvpr,endres14pami,dai15cvpr}.
(2) Size:
Things occur at characteristic sizes with little variance, whereas stuff regions are highly variable in size~\cite{forsyth96orcv,adelson01spie,heitz08eccv}.
(3) Parts:
Thing classes have identifiable parts~\cite{wang15cvpr,felzenszwalb10pami}, 
whereas stuff classes do not (e.g. a \emph{piece of grass} is still \emph{grass}, but a \emph{wheel} is not a \emph{car}).
(4) Instances:
Stuff classes are typically not countable~\cite{adelson01spie} and have no clearly defined instances~\cite{dai15cvpr,hariharan14eccv,tighe14cvpr}.
(5) Texture:
Stuff classes are typically highly textured~\cite{forsyth96orcv,heitz08eccv,tighe13cvpr,dai15cvpr}.
Finally, a few classes can be interpreted as both stuff and things, depending on the image conditions~(e.g. a large number of \emph{people} is sometimes considered a \emph{crowd}).

Several works have shown that different techniques are required for the detection of stuff and things~\cite{tighe13cvpr,tighe14cvpr,kim12eccv,dai15cvpr}.
Moreover, several works have shown that stuff is a useful contextual cue to detect things and vice versa~\cite{rabinovich07iccv,heitz08eccv,kim12eccv,mottaghi14cvpr,shi17iccv}. 

\begin{table}
\small
\centering
\vspace{-1mm}
\tabresize{
\begin{tabular}{ | L{29mm} | C{12mm} | C{10mm} | C{9mm} | C{9mm} | C{7mm} | }
\hline
\textbf{Dataset}						& \textbf{Images} 	&  \textbf{Classes}	&\textbf{Stuff classes} & \textbf{Thing classes} & \textbf{Year} \\
\hline
\hline
MSRC 21~\cite{shotton06eccv}			&	591		&	21		&	6		&	15	&	2006	\\
KITTI~\cite{geiger12cvpr}				&	203		&	14		&	9		&	4	&	2012	\\
CamVid~\cite{brostow09prl}			&	700		&	32		&	13		&	15	&	2008	\\
Cityscapes~\cite{cordts16cvpr}			&	25,000	&	30		&	13		&	14	&	2016	\\
SIFT Flow~\cite{liu11pami}				&	2,688	&	33		&	15		&	18	& 	2009	\\
Barcelona~\cite{tighe10eccv}			&	15,150	&  	170		&	31		&	139	&	2010	\\
LM+SUN~\cite{tighe13ijcv}				&	45,676	&  	232		&	52		&	180	&	2010	\\
PASCAL Context~\cite{mottaghi14cvpr}	&	10,103	& 	540		&	152		& 	388	&	2014	\\
NYUD~\cite{silberman12eccv}			&	1,449	&	894		&	190		&	695	&	2012	\\
ADE20K~\cite{zhou17cvpr}				&	25,210	& 	2,693	& 	1,242 	&   1,451 	&	2017	\\
\hline
COCO-Stuff							&	163,957	& 	172		& 	91		& 	80	&	2018	\\
\hline
\end{tabular}
}
\tabcapstartspace
\vspace{0mm}
\caption{
\textit{
\small
An overview of datasets with pixel-level stuff and thing annotations.
COCO-Stuff is the largest existing dataset with dense stuff and thing annotations.
The number of stuff and thing classes are estimated given the definitions in Sec.~\ref{sec:relatedwork}.
Sec.~\ref{sec:dataset_comparison} shows that COCO-Stuff also has more \emph{usable} classes than any other dataset.
}
}
\tabcapendspace
\vspace{-1mm}
\label{tab:datasets}
\end{table}

\vspace{0mm}
\mypar{Stuff-only datasets.}
Early stuff datasets~\cite{brodatz66doverpublications,dana99tog,lazebnik05pami,caputo10imviscomp} focused on texture classification and had simple images completely covered with a single textured patch.
The more recent Describable Textures Dataset~\cite{cimpoi14cvpr} instead collects textured patches in the wild, described by human-centric attributes.
A related task is material recognition~\cite{sharan14jv,bell13siggraph,bell15cvpr}.
Although the recent Materials in Context dataset~\cite{bell15cvpr} features realistic and difficult images, they are mostly restricted to indoor scenes with man-made materials.
For the task of semantic segmentation,
the Stanford Background dataset~\cite{gould09iccv} offers pixel-level annotations for seven common stuff categories and a single \emph{foreground} category (confounding all thing classes).
All stuff-only datasets above have no distinct thing classes, which make them inadequate to study the relations between stuff an thing classes.

\vspace{-0mm}
\mypar{Thing-only datasets.}
These datasets have bounding box or outline-level annotations of things, e.g. PASCAL VOC~\cite{everingham15ijcv}, ILSVRC~\cite{russakovsky15ijcv}, COCO~\cite{lin14eccv}.
They have pushed the state-of-the-art in Computer Vision, but the lack of stuff annotations limits the ability to understand the whole scene.

\vspace{-0mm}
\mypar{Stuff and thing datasets.}
Some datasets have pixel-wise stuff and thing annotations (Table~\ref{tab:datasets}).
Early datasets like MSRC 21~\cite{shotton06eccv}, NYUD~\cite{silberman12eccv}, CamVid~\cite{brostow09prl} and SIFT Flow~\cite{liu11pami} annotate less than 50 classes on less than 5,000 images.
More recent large-scale datasets like Barcelona~\cite{tighe10eccv}, LM+SUN~\cite{tighe13ijcv}, PASCAL Context~\cite{mottaghi14cvpr}, Cityscapes~\cite{cordts16cvpr} and ADE20K~\cite{zhou17cvpr} annotate tens of thousands of images with hundreds of classes.
We compare COCO-Stuff to these datasets in Sec.~\ref{sec:dataset_comparison}.

\vspace{-0mm}
\mypar{Annotating datasets.}
Dense pixel-wise annotation of images is extremely costly.
Several works use interactive segmentation methods~\cite{rother04siggraph,wang14cviu,castrejon17cvpr} to speedup annotation;
others annotate superpixels~\cite{yamaguchi12cvpr,galasso12accv,ponttuset15cbmi}.
Some works operate in a weakly supervised scenario,
deriving full image annotations starting from manually annotated squiggles~\cite{bearman16eccv,xu15cvpr} or points~\cite{bearman16eccv,jain16hcomp}.
These approaches take less time, but typically lead to lower quality.

In this work we introduce a new annotation protocol to obtain high quality pixel-wise stuff annotations at low human costs
by using superpixels and by exploiting the existing detailed thing annotations of COCO~\cite{lin14eccv} (Sec.~\ref{sec:protocol}).
\presectionspace
\vspace{-5mm}
\section{The COCO-Stuff dataset}
\label{sec:dataset}
\vspace{-1mm}

The Common Objects in COntext~(COCO)~\cite{lin14eccv} dataset is a large-scale dataset of images of high complexity.
COCO has been designed to enable the study of thing-thing interactions,  
and features images of complex scenes with many small objects, annotated with very detailed outlines.
However, COCO is missing stuff annotations.
In this paper we augment COCO by adding dense pixel-wise stuff annotations.
Since COCO is about complex, yet natural scenes containing substantial areas of stuff,
COCO-Stuff enables the exploration of rich relations between things and stuff.
Therefore COCO-Stuff offers a valuable stepping stone towards complete scene understanding.

\begin{figure*}
\small
\centering
\includegraphics[width=1\textwidth]{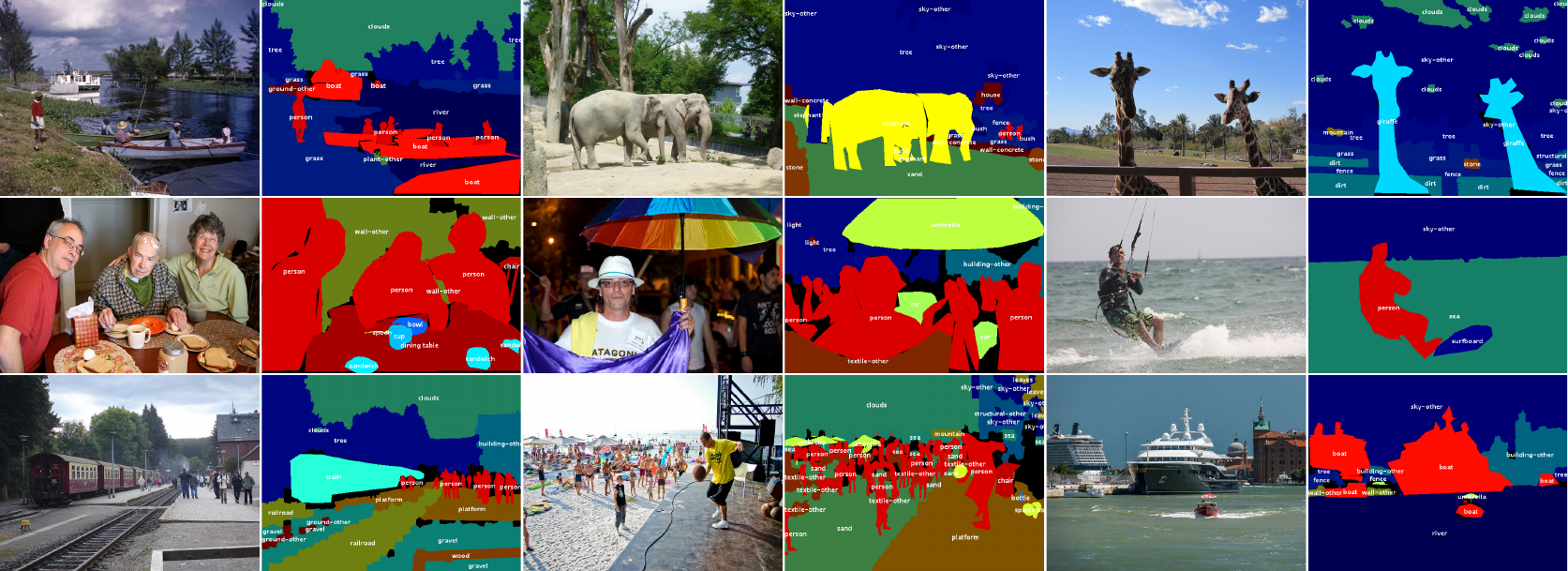}
\figcapstartspace
\vspace{-1mm}
\caption{
\textit{
\small
Annotated images from the COCO-Stuff dataset with dense pixel-level annotations for stuff and things.
To emphasize the depth ordering of stuff and thing classes we use bright colors for thing classes and darker colors for stuff classes.
}
}
\figcapendspace
\vspace{-1mm}
\label{fig:imageexamples}
\end{figure*}

Fig.~\ref{fig:imageexamples} presents several annotated images from the COCO-Stuff dataset, showcasing the complexity of the images,
the large number and diversity of stuff classes, the high level of accuracy of the annotations, and the completeness in terms of surface coverage of the annotations.
We have annotated all 164K images in COCO 2017: training (118K), val (5K), test-dev (20K) and test-challenge (20K).

\presubsectionspace
\subsection{Defining stuff labels.}
\label{sec:definelabels}
COCO-Stuff contains 172 classes: 80 thing, 91 stuff, and 1 class \emph{unlabeled}.
The 80 thing classes are the same as in COCO~\cite{lin14eccv}.
The 91 stuff classes are curated by an expert annotator.
The class \emph{unlabeled} is used in two situations:
if a label does not belong to any of the 171 predefined classes,
or if the annotator cannot infer the label of a pixel.

Before annotation, we choose to predefine our label set.
This contrasts with a common choice in semantic segmentation to have annotators use free-form text labels~\cite{tighe10eccv,tighe13ijcv,mottaghi14cvpr}.
However, using free-form labels leads to several problems. 
First of all, it leads to an extremely large number of classes, many having only a handful of examples.
This makes most classes unusable for recognition purposes, as observed in~\cite{mottaghi14cvpr,zhou17cvpr}.
Furthermore, different annotators typically use several synonyms to indicate the same class.
These need to be merged a posteriori~\cite{tighe10eccv,xiao14ijcv}.
Even after merging, classes might not be consistently annotated.
For example, PASCAL Context~\cite{mottaghi14cvpr} includes the classes \emph{bridge} and \emph{footbridge}, which are in a parent-child relationship.
If one image has \emph{bridge} annotations and another image has \emph{footbridge} annotations,
both can describe the same concept~(i.e. \emph{footbridge}),
or the \emph{bridge} can be another type of \emph{bridge} and therefore describe a different concept.
Similarly, in SIFT Flow~\cite{liu11pami} some images have \emph{field} annotations,
whereas others have \emph{grass} annotations. 
These concepts are semantically overlapping, but are neither synonymous nor in a parent-child relationship. 
A region with a \emph{grass field} could be annotated as \emph{grass} or as \emph{field} depending on the annotator.

\begin{figure}
\small
\vspace{-3mm}
\centering
\includegraphics[width=0.891\linewidth]{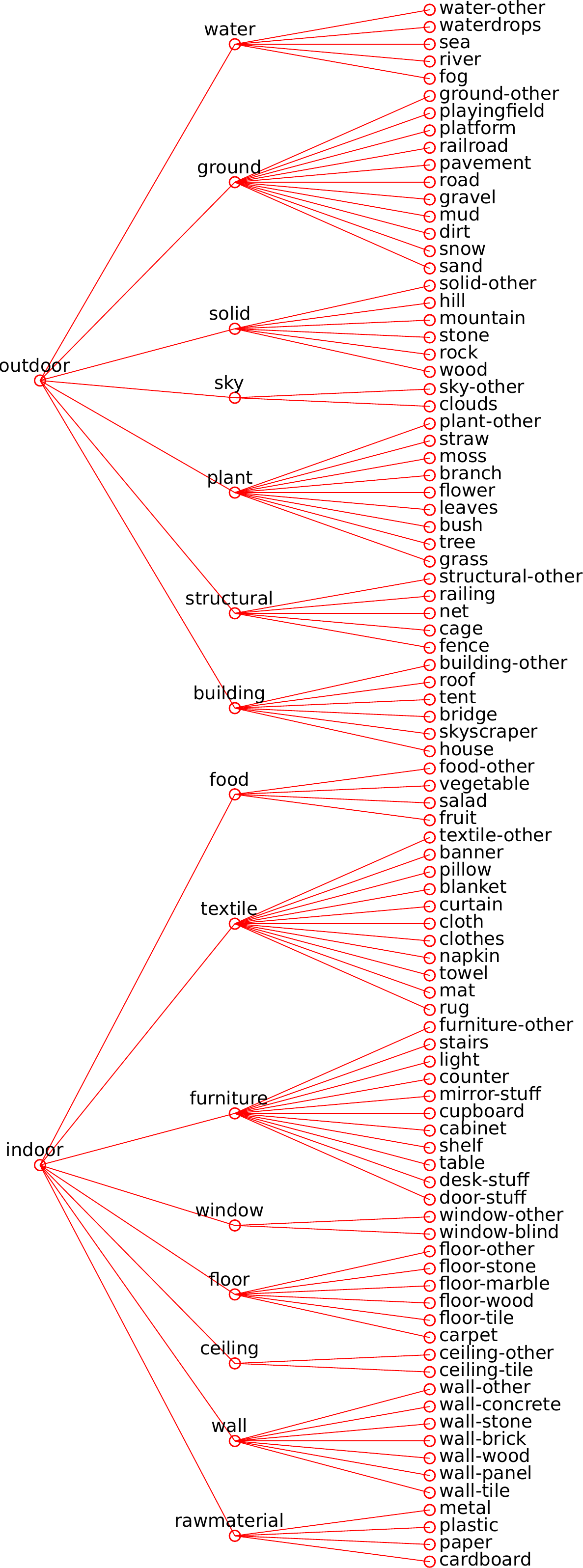}
\figcapstartspace
\vspace{+2mm}
\caption{
\textit{
\small
The stuff label hierarchy of the COCO-Stuff dataset.
Stuff classes are divided into \emph{outdoor} and \emph{indoor}, each further divided into super-categories~(e.g. \emph{floor},
\emph{plant}), and finally into leaf-level classes (e.g. \emph{marble floor, grass}).
The labels used by the annotators form the leaf nodes of the tree.
Furniture classes can be interpreted as either things or stuff, depending on the imaging conditions.
A full list of descriptions is available \href{https://github.com/nightrome/cocostuff/blob/master/labels.md}{online}\textsuperscript{\ref{dataseturl}}.
}
}
\label{fig:labelhierarchy}
\end{figure}

To prevent such inconsistencies, we decided to predefine a set of mutually exclusive stuff classes,
similarly to how the COCO thing classes were defined.
Additionally, we organized our classes into a label hierarchy, 
e.g. classes like \emph{cloth} and \emph{curtain} have \emph{textile} as parent,
while classes like \emph{moss} and \emph{tree} have \emph{vegetation} as parent (Fig.~\ref{fig:labelhierarchy}).
The super-categories \emph{textile} and \emph{vegetation} have \emph{indoor} and \emph{outdoor} as parents, respectively.
The top-level nodes in our hierarchy are generic classes \emph{stuff} and \emph{thing}.

To choose our set of stuff labels, the expert annotator used the following criteria:
stuff classes should
(1) be mutually exclusive;
(2) in their ensemble, cover the vast majority of the stuff surface appearing in the dataset;
(3) be frequent enough;
(4) have a good level of granularity, around the base level for a human.
However, these criteria conflict with each other:
if we label all vegetations as \emph{vegetation}, the labels are too general.
On the other extreme, if we create a separate class for every single type of \emph{vegetation}, the labels are too specific and infrequent.
Therefore, as shown in Fig.~\ref{fig:labelhierarchy}, for every super-category like \emph{vegetation},
we explicitly list its most frequent subclasses as choices for the annotator to pick~(e.g. \emph{straw}, \emph{moss}, \emph{bush} and \emph{grass}).
And there is one additional subclass \emph{vegetation-other} to be picked to label any other case of \emph{vegetation}.
This achieves the coverage goal, while avoiding to scatter the data over many small classes.
For some super-categories~(\emph{floor}, \emph{wall} and \emph{ceiling}) we are particularly interested in the material they are made of.
Therefore we include the material type in the class definition~(e.g. \emph{wall-brick}, \emph{wall-concrete} and \emph{wall-wood}).
This enables further analysis of the materials present in a scene.

\begin{figure}
\small
\vspace{0mm}
\centering
\includegraphics[width=1\linewidth]{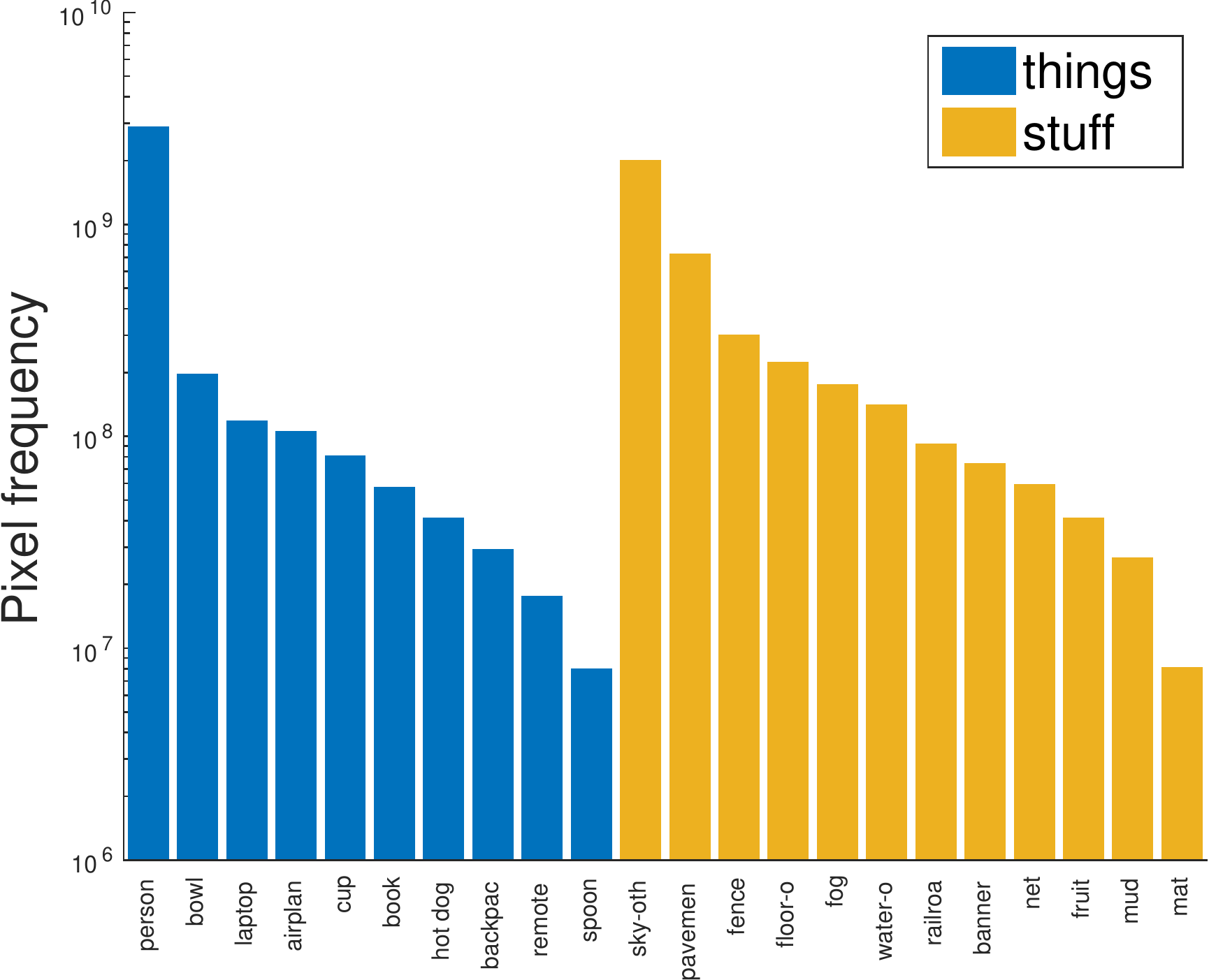}
\figcapstartspace
\vspace{-2mm}
\caption{
\textit{
\small
Pixel-level frequencies for some of the classes in the trainval set of COCO-Stuff. 
For clarity, we show about $1/8$ of all classes.
We can see that stuff and thing classes follow a similar pixel frequency distribution.
}
}
\figcapendspace
\vspace{+2mm}
\label{fig:labelfreqs}
\end{figure}

Our label set fulfills all design criteria~(1-4):
(1) the mutual exclusivity of labels is by design and enforced through having annotators only use the leaves of our hierarchy as labels~(Fig.~\ref{fig:labelhierarchy}).
For the other criteria we need to look at pixel-level frequencies after dataset collection:
(2) only 6\% of the pixels are \emph{unlabeled}, which is satisfactory;
(3, 4) interestingly, all our stuff classes have pixel frequencies in the same range of the COCO thing classes~(Fig.~\ref{fig:labelfreqs}) and they also follow a similar distribution and granularity~(Fig.~\ref{fig:labelhierarchy}).
Intuitively, having both thing and stuff classes follow similar distributions makes the dataset well suited to analyze stuff-thing relations.

\begin{figure}
\small
\vspace{-0mm}
\centering
\includegraphics[width=1\linewidth]{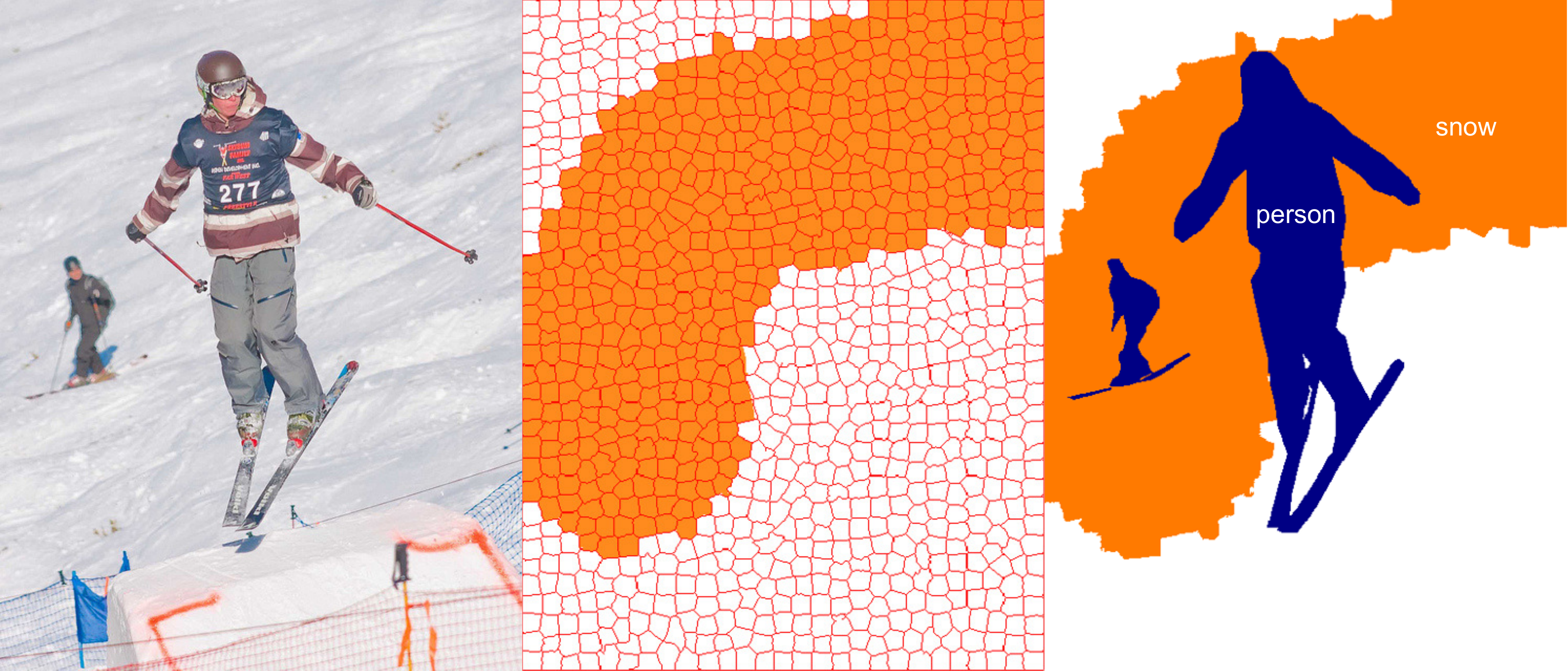}
\figcapstartspace
\vspace{-0mm}
\caption{
\textit{
\small
Example of a) an image, b) the superpixel-based stuff annotation and c) the final labeling.
The annotator can quickly annotate large stuff regions (\emph{snow}) with a single mouse stroke using a paintbrush tool.
Thing (\emph{person}) annotations are copied from the COCO dataset.
The transparency of each layer can be regulated to get a better overview.
This approach dramatically reduces annotation time and yields a very accurate labeling, especially at stuff-thing boundaries.
}
}
\figcapendspace
\vspace{+2mm}
\label{fig:annotationexample}
\end{figure}

\presubsectionspace
\subsection{Annotation protocol and analysis}
\label{sec:protocol}

\mypartop{Protocol.}
We developed a very efficient protocol, specialized for labeling stuff classes at the pixel-level.
We first partition each image into 1,000 superpixels using SLICO~\cite{achanta12pami},
which adheres very well to boundaries and gives superpixels of homogeneous size~(Fig.~\ref{fig:annotationexample}).
Superpixels remove the need for manually delineating the exact boundaries between two regions of different classes.
As superpixels respect boundaries, it is enough to mark which superpixels belong to which class, which is a lot faster to do.
Moreover, the evenly spaced and sized SLICO superpixels result in a labeling task natural for humans
(as opposed to superpixel algorithms which yield regions that greatly vary in size~\cite{felzenszwalb04ijcv}).
We accelerate the annotation process by providing annotators a size-adjustable paintbrush tool,
which enables labeling large regions of stuff very efficiently~(Fig.~\ref{fig:annotationexample}b).

We improve annotation efficiency even further by leveraging the highly accurate thing outlines available from COCO~\cite{lin14eccv} (Fig.~\ref{fig:annotationexample}c).
We show annotators images with \emph{thing overlays}, and pixels belonging to things are clamped and unaffected by the annotator's brush.
This results in a lightweight experience, where the annotator merely needs to select a stuff class (like \emph{snow}) and brush over the foreground object.
In fact, because of the high annotation accuracy of COCO things, our technique results in extremely precise stuff outlines at stuff-thing boundaries, often beyond the accuracy of superpixel boundaries.

As a final element in our protocol, we present our stuff labels to the annotators using the full hierarchy.
In initial trials we found that, compared to presenting them in a list, this reduces the look-up time of labels significantly.
This annotation protocol yields an annotation time of only three minutes to annotate stuff in one of the COCO images, which are very complex~(Fig.~\ref{fig:imageexamples}).
We release the superpixels and the annotation tool online to allow for further analysis.

We annotated 10K images with our protocol using in-house annotators.
Afterwards, we collaborated with the startup Mighty AI to adapt our protocol for crowdsourcing and annotate all remaining images of COCO-Stuff.

\vspace{-0mm}
\mypar{Analysis of superpixels.}
We study here the quality-speed trade-off of using superpixels.
We ask a single annotator to annnotate 10 COCO images three times, once for each of three different modalities:
(1) superpixel annotation, as we do for COCO-Stuff;
(2) polygon annotation, the de facto standard~\cite{cordts16cvpr,mottaghi14cvpr,zhou17cvpr} and
(3) freedraw annotation, which consists of directly annotating pixels with a very accurate size-adjustable paintbrush tool, but \emph{without} aid from superpixels.
The freedraw annotations attempt to get as close to pixel-level accuracy as possible, and we use them as ground-truth reference in this analysis.

Table~\ref{tab:modalities} shows the results for superpixel, polygon and freedraw annotation. 
Compared to the freedraw reference, polygons and superpixels are much faster (1.5x and 2.8x).
Computing pixel-level labeling agreement w.r.t. freedraw reveals that both polygons and superpixels lead to very accurate annotations (96\%-97\%).
We also asked the annotator to re-annotate the images with the same modality, enabling to measure `self agreement'.
Interestingly the self agreement of freedraw is in the same range as the agreement of superpixels and polygons w.r.t. freedraw.
This shows that the differences across annotation modalities are of similar magnitude to the natural variations within a single modality, even by a single annotator.
Hence, all three modalities are about as accurate.

Furthermore we simulate our stuff annotation protocol on two other datasets which were originally annotated with polygons: SIFT Flow~\cite{liu11pami} and PASCAL Context~\cite{mottaghi14cvpr}.
For each image we label each superpixel with the majority stuff label in the ground-truth annotations.
We then overlay the existing thing annotations.
This protocol achieves 98.3\% agreement with the ground-truth on SIFT Flow and 98.4\% on PASCAL Context.
These findings show that superpixel annotation is faster than conventional polygon annotation, while providing almost the same annotations.

\begin{table}
\small
\centering
  \begin{tabular}{ | L{1.8cm} | C{1.4cm} | C{1.6cm} | C{1.6cm} | }
\hline
\textbf{Modality}	&	\textbf{Speedup}	&	\textbf{Reference agreement}	&	\textbf{Self agreement}	\\
\hline
\hline
Superpixels		&	2.8				&	96.1\%						&	98.7\%					\\	
Polygons			&	1.5				&	97.3\%						&	97.0\%					\\
Freedraw		&	1.0				&	-							&	96.6\%					\\
\hline
\end{tabular}
\tabcapstartspace
\caption{
\textit{
\small
A quantitative comparison of different stuff annotation modalities.
We use freedraw annotation as a reference in the 'Speedup' and 'Reference agreement' columns.
The self-agreement between repeated runs of the same annotation modality decreases with weaker constraints on the possible labelings.
}
}
\label{tab:modalities}
\tabcapendspace
\vspace{+2.4mm}
\end{table}

\begin{figure}
\small
\centering
\vspace{-1mm}
\includegraphics[width=\linewidth]{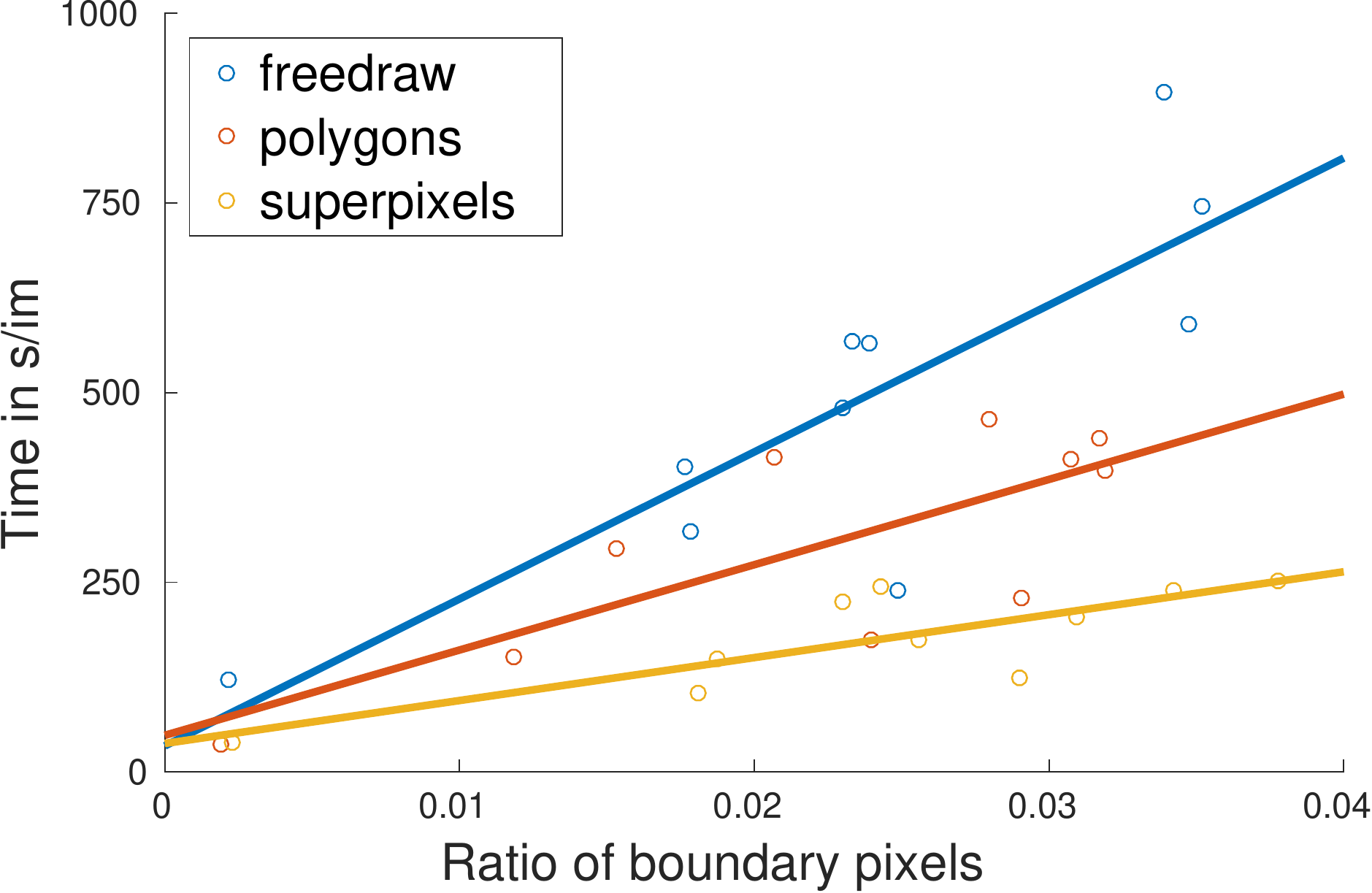}
\figcapstartspace
\vspace{+0.1mm}
\caption{
\textit{
\small
Annotation time versus image boundary complexity.
Each circle indicates an image annotated using one of three modalities.
The trendlines show that annotation time for some mo\-dalities increases faster with boundary complexity than for others.
}
}
\figcapendspace
\vspace{-5.0mm}
\label{fig:boundarycomplexity}
\end{figure}

We found that a dominant factor for the differences in annotation time across images is their boundary complexity.
Boundary complexity is defined as the ratio of pixels that have any neighboring pixel with a different semantic label~(as in the boundary evaluation in~\cite{caesar16eccv,kohli09ijcv,kraehenbuehl11nips}).
Fig.~\ref{fig:boundarycomplexity} analyzes the relationship between boundary complexity and annotation time of an image using different annotation modalities.
The linear trendlines show that there is a clear correlation between annotation time and boundary complexity.
We can see that the slopes of the freedraw and polygon annotation trendlines are 3.4x and 2.0x steeper than for superpixels.
This is one of the main reasons why superpixels yield such big improvements in annotation time on average.

\newpage
\vspace{-0mm}
\mypar{Analysis of thing overlays.}
We analyze thing overlays in terms of the annotation speedup they bring and the quality they lead to.
For this we perform superpixel and freedraw annotation with and without thing overlays.
We achieve significant speedups when using thing overlays with freedraw annotation (1.8x) and also with superpixel annotation (1.2x).
Furthermore, the agreement of superpixel annotation w.r.t. the freedraw reference is identical with and without thing overlays (96.1\% in both cases).
This shows that thing overlays achieve a significant speedup without any loss in quality.

Moreover, 46.8\% of the boundary pixels in COCO-Stuff have a neighboring pixel that belongs to a thing class.
Therefore using thing overlays significantly decreases the boundary complexity and leads to a larger speedup for freedraw annotation than for superpixel annotation.

\mypar{Across-annotator agreement.}
Following~\cite{zhou17cvpr,cordts16cvpr} we annotate 30 images by 3 annotators each.
For each image we compute the label agreement between each pair of annotators and average over all pairs.
The mean label agreement in COCO-Stuff is 73.6\%, compared to 66.8\% for ADE20K~\cite{zhou17cvpr}.

\presubsectionspace
\subsection{Comparison to other datasets.}
\label{sec:dataset_comparison}
COCO-Stuff has the largest number of images of any semantic segmentation dataset (164K).
In particular, MSRC 21~\cite{shotton06eccv}, KITTI~\cite{geiger12cvpr}, CamVid~\cite{brostow09prl}, SIFT Flow~\cite{liu11pami} and NYUD~\cite{silberman12eccv} all have less than 5,000 images~(Table~\ref{tab:datasets}).
COCO-Stuff is also much richer in both the number of stuff and thing classes than MSRC 21~\cite{shotton06eccv}, 
KITTI~\cite{geiger12cvpr}, CamVid~\cite{brostow09prl}, Cityscapes~\cite{cordts16cvpr} and SIFT Flow~\cite{liu11pami}.
Compared to the Barcelona~\cite{tighe10eccv} and LM+SUN~\cite{tighe13ijcv} datasets, it has $3\times$ and $2\times$ more stuff classes, respectively.

\begin{figure}
\small
\centering
\vspace{-1mm}
\includegraphics[width=\linewidth]{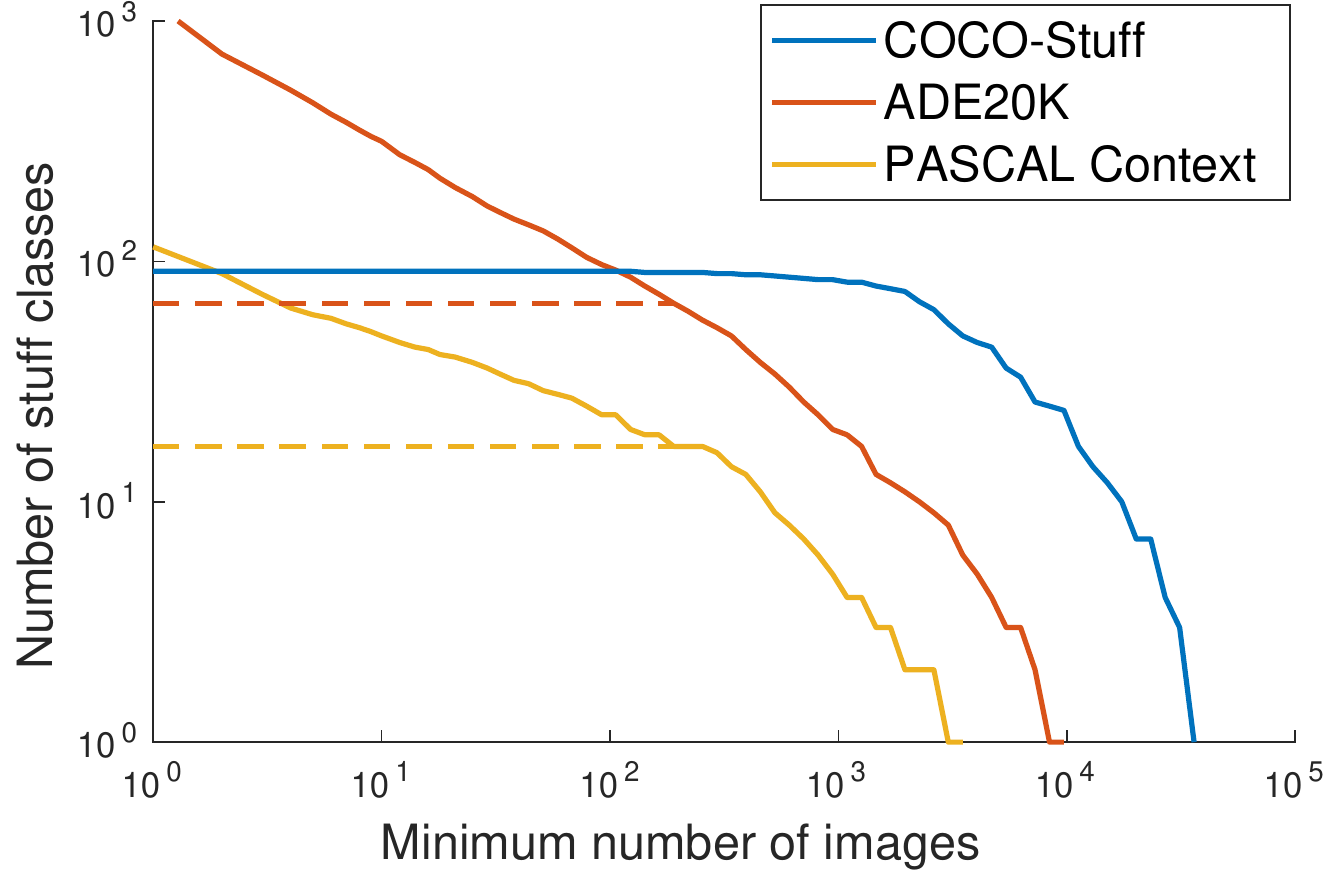}
\figcapstartspace
\vspace{-0.6mm}
\caption{
\textit{
\small
The number of stuff classes occurring in at least $x$ images for varying thresholds of $x$.
Solid lines indicate the full datasets, dashed lines the versions with only usable classes.
Statistics are computed on the trainval sets of three datasets.
}
}
\figcapendspace
\vspace{-0.85mm}
\label{fig:datasetcomparison}
\end{figure}

PASCAL Context~\cite{mottaghi14cvpr} and ADE20K~\cite{zhou17cvpr} are the most similar datasets to COCO-Stuff.
On the surface they appear to have a very large numbers of classes (540 and 2,693), but in practice most classes are rare.
The authors of those datasets define a set of classes deemed \emph{usable} for experiments (i.e. the most frequent 60 classes in PASCAL Context and 150 classes in ADE20K).
In Fig.~\ref{fig:datasetcomparison} we show the number of stuff classes that occur in at least $x$ images, for varying thresholds $x$, on the trainval sets of three datasets.
COCO-Stuff has more usable stuff classes than PASCAL Context and ADE20K for \emph{any} threshold,
e.g. for $x = 1,000$, there are 5 stuff classes in PASCAL Context, 20 in ADE20K and 84 in COCO-Stuff.
This means that 92\% of the stuff classes in COCO-Stuff occur in at least $1,000$ images.
Furthermore, both PASCAL Context and ADE20K use free-form label names, 
which lead to annotations at different granularities and hence ambiguities, as discussed in Sec.~\ref{sec:definelabels}.
In contrast, in COCO-Stuff all labels are mutually exclusive and at a comparable level of granularity.
Finally, PASCAL Context and ADE20K are annotated with overlapping polygons.
Hence some pixels have multiple conflicting labels at the boundaries between things and stuff.
In COCO-Stuff instead, each pixel has exactly one label.

To conclude, COCO-Stuff is a very large dataset of highly complex images.
It has the largest number of usable stuff and thing classes with pixel-level annotations.
Moreover, by building on COCO it also has natural language captions, further supporting rich scene understanding.

\presectionspace
\vspace{+5mm}
\section{Analysis of stuff and things}
\label{sec:analysis}
\vspace{-0mm}

In this section we leverage COCO-Stuff to analyze various relations between stuff and things:
we analyze the relative importance of stuff and thing classes~(Sec.~\ref{sec:importance});
study spatial contextual relations between stuff and things~(Sec.~\ref{sec:context});
and analyze the behavior of semantic segmentation methods on stuff and things~(Sec.~\ref{sec:semsegm}).
To preserve the integrity of the test set annotations, all experiments in this section are run on the trainval set of COCO-Stuff.

\newpage
\presubsectionspace
\vspace{-0mm}
\subsection{Importance of stuff and things}
\label{sec:importance}
\vspace{-0.0mm}
We quantify the relative importance of stuff and things using two criteria: surface cover and human descriptions.

\vspace{-1.0mm}
\mypar{Surface cover.}
We measure the frequencies of stuff and thing pixels in the COCO-Stuff annotations.
Table~\ref{tab:importance} shows that the majority of pixels are stuff (69.1\%).
We also compute statistics for the labeled \emph{regions} in COCO-Stuff,
i.e. connected components in the pixel annotation map.
We use such regions as a proxy for class instances, as stuff classes do not have instances.
We see that 69.4\% of the regions are stuff and 30.6\% things.

\vspace{-1mm}
\mypar{Human descriptions.}
Although stuff classes cover the majority of the image surface, one might argue they are just irrelevant background pixels. 
The COCO dataset is annotated with five captions per image~\cite{lin14eccv},
which have been written explicitly to describe its content, and therefore capture the most relevant aspects of the image for a human.
To emphasize the importance of stuff for scene understanding, we also analyze these captions, counting how many nouns point to things and stuff respectively.
We use a Part-Of-Speech~(POS) tagger~\cite{toutanova03naacl} to automatically detect nouns.
Then we manually categorize the 600 most frequent nouns as stuff (e.g. \emph{street}, \emph{field}, \emph{water}, \emph{building}, \emph{beach}) or things (e.g. \emph{man}, \emph{dog}, \emph{train}),
ignoring nouns that do not represent physical entities (e.g. \emph{game}, \emph{view}, \emph{day}).

Table~\ref{tab:importance} shows the relative frequency of these nouns.
Stuff covers more than a third of the nouns (38.2\%).
This clearly shows the importance of stuff according to the COCO image captions.

\begin{table}
\small
\centering
 \begin{tabular}{ | L{2.5cm} | C{2.2cm} | C{2.2cm} | }
\hline
\textbf{Level}		&	\textbf{Stuff}		&	\textbf{Things}	\\
\hline
\hline
Pixels			&	\textbf{69.1\%}	&	30.9\%			\\ 
Regions			&	\textbf{69.4\%}	&	30.6\%			\\
\hline
Caption nouns	&	38.2\%			&	\textbf{61.8\%}	\\
\hline
\end{tabular}
\tabcapstartspace
\vspace{-1.0mm}
\caption{
\textit{
\small
Relative frequency of stuff and thing classes in pixel-level annotations and caption nouns in COCO-Stuff.
}
}
\label{tab:importance}
\tabcapendspace
\vspace{-2mm}
\end{table}

\presubsectionspace
\subsection{Spatial context between stuff and things}
\label{sec:context}
\vspace{-0mm}

\begin{figure*}
\centering
\small
\includegraphics[width=\linewidth]{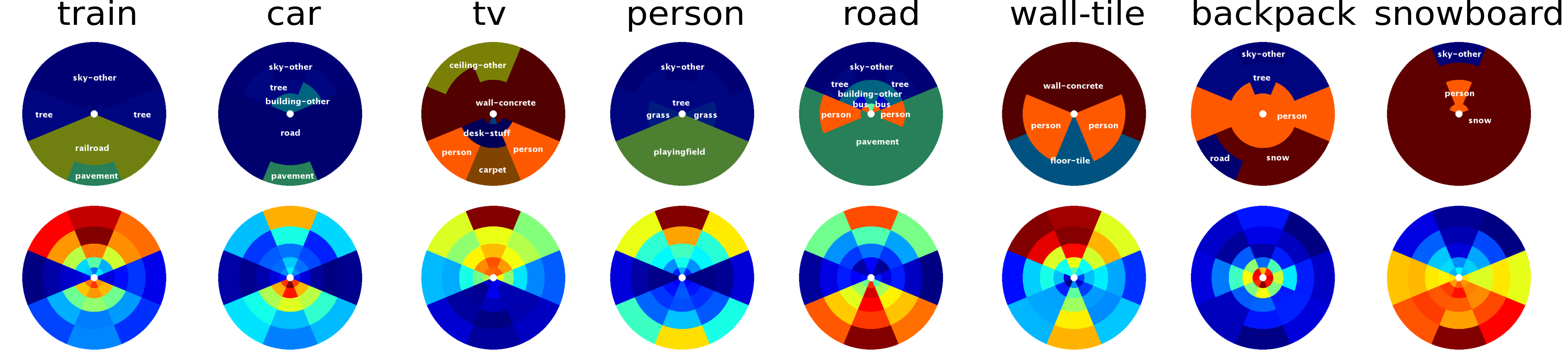}
\figcapstartspace
\caption{
\textit{
\small
Spatial context visualizations.
(Top) Each disc is for a different reference class and shows the most likely other class at each direction and distance bin.
(Bottom) The conditional probabilities of the most common class in each bin, as a measure of confidence.
The values are normalized for each reference class and range from low~(blue) to high~(red).
We also show examples for classes with high~(person) and low~(snowboard) mean entropy.
}
}
\figcapendspace
\label{fig:spatialcontext}
\end{figure*}

\mypartop{Methodology.}
We analyze spatial context by considering the relative image position of one class with respect to another.
For simplicity, here we explain how to compute the spatial context for one particular reference class,
i.e. \emph{car} (Fig.~\ref{fig:spatialcontext}, second column).
The explanation is analogous for all other classes.
For every image containing a \emph{car}, we extract a set of \emph{car} regions, i.e. connected components of \emph{car} pixels in the annotation map.
Next we compute a histogram of image pixels surrounding the \emph{car} regions, with two spatial dimensions~(distance, angle) and one dimension for the class label.
To determine in which spatial bin a certain pixel lands, we
(1) compute the distance between the pixel and the nearest point in the \emph{car} region (normalized by image size);
(2) compute the relative angle with respect to the center of mass of the \emph{car} region.

\mypar{Results.}
Fig.~\ref{fig:spatialcontext} shows the spatial context of eight reference classes.
This visualization reveals several interesting contextual relations.
\emph{Trains} are typically found above \emph{railroads}~(thing-stuff).
\emph{TVs} are typically found in front of \emph{persons}~(thing-thing).
\emph{Tiled walls} occur above \emph{tiled floors}~(stuff-stuff), and \emph{roads} are flanked by \emph{persons} on both sides~(stuff-thing).
Note that these contextual relations are not necessarily symmetric: most \emph{cars} appear above a \emph{road}, but many \emph{roads} have other things above them.

For each reference class and spatial bin we also show the conditional probability of the most likely other class as a measure of confidence~(Fig.~\ref{fig:spatialcontext}, bottom).
In most cases the highest confidence is in regions above (\emph{sky}, \emph{wall}, \emph{ceiling}) or below (\emph{road}, \emph{pavement}, \emph{snow}) the reference region,
but rarely to the left or right.
Since vertical relations are mostly support relations (e.g. `on top of'), this suggests that support is the most informative type of context.
For some classes the highest confidence region is also very close to the reference region, often indicating that another object is attached to the reference one (\emph{person} close to \emph{backpack}).

As the figure shows, some classes have a rich and diverse context, composed of many other classes (e.g. \emph{tv}, \emph{road}),
while some classes have a simpler context (e.g. \emph{snowboards} always appear in the middle of \emph{snow}).
We quantify the complexity of a reference class as the entropy of the conditional probability distribution, averaged over all other classes and spatial bins.
The classes with highest mean entropy are \emph{wood}, \emph{metal} and \emph{person}, and those with the lowest are \emph{snowboard}, \emph{airplane} and \emph{playingfield}.
On average, we find that stuff classes have a significantly higher mean entropy than things~(3.40 vs. 3.02), showing they appear in more varied contexts.
We also find that the mean entropy is rather constant over distances~(small: 3.21, big: 3.23) and directions~(left: 3.19, right: 3.18, down: 3.20, up: 3.15).

Comparing the mean entropy of different datasets, taking into account all classes,
we find that COCO-Stuff has the highest~(3.22), followed by the 60 \emph{usable} classes of PASCAL Context~(2.42),
the 150 \emph{usable} classes of ADE20K~(2.18) and SIFT Flow~(1.20).
This shows the contextual richness of COCO-Stuff.

\presubsectionspace
\subsection{Semantic segmentation of stuff and things}
\label{sec:semsegm}
\vspace{-1mm}
We now analyze how a modern semantic segmentation method~\cite{chen15iclr} performs on COCO-Stuff.
We compare the performance on stuff and thing classes
and hope to establish a baseline for future experiments on this dataset.

\mypar{Protocol.}
We use the popular DeepLab V2~\cite{chen15iclr} based on the VGG-16 network~\cite{simonyan15iclr} pre-trained on the ILSVRC classification dataset~\cite{russakovsky15ijcv}.
We use the following experimental protocol: 
train on the 118K training images and test on the 5K val images.
To evaluate performance we use four criteria commonly used in the literature~\cite{long15cvpr,eigen15iccv,caesar16eccv}:
(1) {\em pixel accuracy} is the percentage of correctly labeled pixels in the dataset,
(2) {\em class accuracy} computes the average of the per-class accuracies,
(3) {\em mean Intersection-over-Union~(IOU)} divides the number of pixels of the intersection of the predicted and ground-truth class by their union, averaged over classes~\cite{everingham15ijcv},
(4) {\em frequency weighted~(FW) IOU} is per-class IOU weighted by the pixel-level frequency of each class.

\vspace{-0.0mm}
\mypar{Results for all images and classes.}
Table~\ref{tab:moredata} shows the results using all images (row ``118K (train)").
DeepLab achieves an mIOU of 33.2\% over all classes.
A detailed comparison of leading methods can be found \href{https://github.com/nightrome/cocostuff}{online}\textsuperscript{\ref{dataseturl}}.

\vspace{-0.0mm}
\mypar{Benefits of a large dataset.}
One reason for the recent success of deep learning methods is the advent of large-scale datasets~\cite{russakovsky15ijcv,ionescu15iccv,zhou17cvpr}.
Inspired by~\cite{sun17iccv}, we want to test whether the performance of semantic segmentation models plateaus at current dataset sizes or whether it benefits from larger datasets.
Following the above protocol, we train multiple DeepLab models with different amounts of training data, keeping all training parameters fixed.
Table~\ref{tab:moredata} shows the resulting performance on the validation set (rows from 1K to 118K).
We can see that for all metrics, performance significantly increases as the training set grows.
We hypothesize that even deeper network architectures~\cite{he16cvpr} could benefit even more from large training sets.

\vspace{-0.0mm}
\mypar{Is stuff easier than things?}
Several works found that stuff is easier to segment than things \cite{tighe10eccv,ion11nips,liu11pami,tighe13cvpr,tighe14cvpr,zhang15icra,xu15cvpr,zhou17cvpr}.
We argue that this is due to their choice of dataset, rather than a general observation.
Most datasets only include a small number of very frequent and coarse-grained stuff classes, such as \emph{sky} and \emph{grass}~(Table~\ref{tab:datasets}).
In contrast, COCO-Stuff features a larger number of relevant stuff labels at a similar level of granularity as the existing thing labels.
It has a similar number of stuff and thing classes, and a similar pixel frequency distribution for both~(see Fig.~\ref{fig:labelfreqs}).

As Table~\ref{tab:moredata} (bottom) shows, on COCO-Stuff DeepLap performs substantially better on thing classes than on stuff.
This shows that stuff is harder to segment than things in COCO-Stuff, 
a dataset where both stuff and things are similarly distributed. 
Therefore we argue that stuff is not generally easier than things.

\begin{table}
\small
\centering
\begin{tabular}{ | C{1.6cm} | C{1.3cm} | C{1.3cm} | C{0.9cm} | C{0.9cm} |}
\hline
 \textbf{Training images}	& \textbf{Class accuracy}&  \textbf{Pixel accuracy} 	& \textbf{Mean IOU}	& \textbf{FW IOU}\\
\hline
\hline
1K						& 24.1\%			& 46.1\%			& 15.9\%			& 31.0\%\\
5K						& 33.8\%			& 52.7\%			& 23.1\%			& 37.5\%\\
10K						& 36.9\%			& 54.6\%			& 25.5\%			& 39.6\%\\
20K						& 40.2\%			& 57.5\%			& 28.6\%			& 42.6\%\\
40K						& 43.0\%			& 61.1\%			& 31.4\%			& 45.7\%\\
80K						& 44.9\%			& 63.4\%			& 32.9\%			& 47.4\%\\
118K (train)				& \textbf{45.1\%}	& \textbf{63.6\%}	& \textbf{33.2\%}	& \textbf{47.6\%}\\
\hline
\hline
stuff					& 33.5\%			& 58.2\%			& 24.0\%			& 45.6\%\\
things					& \textbf{58.3\%}	& \textbf{75.7\%}	& \textbf{43.6\%}	& \textbf{58.4\%}\\
\hline
\end{tabular}
\tabcapstartspace
\caption{
\textit{
\small
Rows 1K to 118K: Performance of Deeplab V2 with VGG-16 with varying amounts of training data.
We can see that for all metrics, performance significantly increases for larger datasets.
Last two rows: Performance of the same model on stuff and thing classes using all 118K training images in COCO.
}
}
\label{tab:moredata}
\tabcapendspace
\vspace{-2mm}
\end{table}

\presectionspace
\vspace{+3mm}
\section{Conclusion}
\label{sec:conclusion}

We introduced the large-scale COCO-Stuff dataset.
COCO-Stuff enriches the COCO dataset with dense pixel-level stuff annotations.
We used a specialized stuff annotation protocol to efficiently label each pixel.
Our dataset features a diverse set of stuff classes.
In combination with the existing thing annotations in COCO it allows us to perform a detailed analysis of stuff and the rich contextual relations that make our dataset unique.
We have shown that
(1) stuff is important: Stuff classes cover the majority of the image surface and more than a third of the nouns in human descriptions of an image;
(2) many classes show frequent patterns of spatial context, and stuff classes appear in more varied contexts than things;
(3) stuff is not generally easier to segment than things;
(4) the larger training set that COCO-Stuff offers improves the semantic segmentation performance.

\vspace{+1mm}
\mypar{Acknowledgments.}
This work is supported by the ERC Starting Grant VisCul.
The annotations were done by the crowdsourcing startup Mighty AI,
and financed by Mighty AI and the Common Visual Data Foundation.

\clearpage
{\small
\bibliographystyle{ieee}
\bibliography{../../../bibtex/shortstrings,../../../bibtex/calvin,../../../bibtex/vggroup}
}
\end{document}